\documentclass[conference]{IEEEtran}
\IEEEoverridecommandlockouts

\usepackage{microtype}
\usepackage{graphicx}
\usepackage{subfigure}
\usepackage{booktabs} 
\usepackage{amsmath,amssymb,amsfonts}

\usepackage{hyperref}

\def\BibTeX{{\rm B\kern-.05em{\sc i\kern-.025em b}\kern-.08em
    T\kern-.1667em\lower.7ex\hbox{E}\kern-.125emX}}
\begin{document}

\title{Channel Locality Block: A Variant of Squeeze-and-Excitation\\}

\author{\IEEEauthorblockN{1\textsuperscript{st} Huayu Li}
\IEEEauthorblockA{\textit{Northern Arizona University}\\
Flagstaff, United State\\
\textit{Northern Arizona University}\\
hl459@nau.edu}
}

\maketitle

\begin{abstract}
Attention mechanism is a hot spot in deep learning field. Using channel attention model is an effective method for improving the performance of the convolutional neural network. Squeeze-and-Excitation \cite{hu2017squeeze} block takes advantage of the channel dependence, selectively emphasizing the important channels and compressing the relatively useless channel. In this paper, we proposed a variant of SE block based on channel locality. Instead of using full connection layers to explore the global channel dependence, we adopt convolutional layers to learn the correlation between the nearby channels. We term this new algorithm Channel Locality(C-Local) block. We evaluate SE block and C-Local block by applying them to different CNNs architectures on cifar-10 dataset. We observed that our C-Local block got higher accuracy than SE block did.
\end{abstract}


\section{Introduction}
\label{Introduction}
In recent years, convolutional neural networks have made significant breakthroughs in many fields, especially in computer vision \cite{krizhevsky2012imagenet} \cite{ren2015faster} \cite{long2015fully} \cite{xie2017aggregated}. The convolution kernel, which is the core of the convolutional neural network, is generally regarded as an information aggregate that aggregates spatial information and channel-wise information in the local receptive field. The convolutional neural network consists of a series of convolutional layers, nonlinear layers, and downsampling layers. As a result, it can capture the features of the image from the global receptive field to describe the images. In general, we can regard the convolutional layer as a set of filters which are learned to express local spatial connectivity patterns along input channels. It means that by combining the spatial and channel-wise information within local receptive fields, the convolutional filters are trained to contain the information about the original input images. Stacking more convolutional layers and increasing the filters contained in the convolutional layers has been proven to be an effective method to enable the CNNs a powerful capability to extract the features of the input. In other words, wider and deeper CNNs have a better performance in most visual tasks. Embedding learning also helps improve the performance of CNNs. Inception \cite{szegedy2015going} architectures, which use multi-branch architectures, show that network can achieve higher accuracy by embedding learning. More recent works show the obvious benefits of another important mechanism, attention mechanisms \cite{jaderberg2015spatial} \cite{hu2017squeeze}, which explore the spatial and channel correlations.

In this paper, we propose a variant of SEnet \cite{hu2017squeeze} which focus on the near channels relationship instead of the whole channels. We term our new architecture the “Channel Locality”(C-Local) block. Through the early work \cite{lecun1995convolutional} \cite{krizhevsky2012imagenet}, the local correlation of the data was found. The convolution operation appeared and replaced most of the work of the full connection so that the model can extract a large number of useful features while ensuring the streamlining of the model parameters. The locality of height and width dimensions is effectively verified by CNNs. Here we throw a question, whether the channel information has locality? In other words, when processing with channel information, can we not take all the information into account like taking processing with the height and width? To prove this, we investigate our algorithm which focuses on the correlations of the nearby area of channels by learning global information to scale the feature maps.

\begin{figure}[ht]
\vskip 0.2in
\begin{center}
\centerline{\includegraphics[width=\columnwidth]{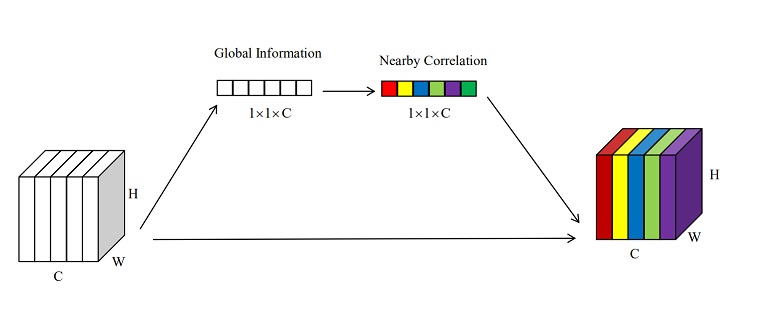}}
\caption{A Channel Locality block. We can simply see that it is a variant of a Squeeze-and-Excitation block. Instead of using full connection layers to learn the global channel correlation, we focus on the correlation between nearby channels.}
\label{fig1}
\end{center}
\vskip -0.2in
\end{figure}

The basic structure of the C-Local block is illustrated in Figure~\ref{fig1}. For any feature extraction module(Generally, a convolutional layer or a set of convolutional layers), a corresponding C-Local block can be applied to perform the channel attention. The features are firstly extracted the global information and followed by a weight layer to capture the nearby channel correlation. The deployment strategy is simply stacking the feature extraction module with C-Local block that enables our algorithm to be easily applied to various existing neural network structures.

\section{Related Work}
\label{Related Work}

\textbf{Local Correlation Of Data:}Before the convolution operation broke out, the main operation of the neural network is the full connections, that is, each element of the output is affected by three dimensions. After the magical effect of the convolution operation was discovered, convolution operations have been extensively used. In the convolution operation, the influence of the three dimensions in each of the output element is as follows: (1) All channels. (2) Partial height. (3) Partial width. Convolution operations occurred and replaced most of the work of full connection, allowing the model to be able to extract a large number of useful features while also ensuring the simplification of model parameters. Therefore, in the two dimensions of height and width, locality is effectively verified. In the work of group convolution\cite{xie2017aggregated}\cite{ma2018shufflenet}\cite{chollet2017xception}, the locality of the channel is verified.

\textbf{Attention Mechanisms:}The visual attention mechanism is a unique brain signal processing mechanism of the human vision system. By quickly scanning the global image, the human vision system obtains the target area that needs to be focused on. And then the vision system invests more attention resources in this area, which called the focus of attention, to obtain more detailed information about the target and suppress other useless information. The attention mechanism in deep learning is essentially similar to human selective visual attention mechanism. The core goal is also to select more information from the many information that is more critical to the current task objectives. The benefits of attention mechanism have been shown in many works  \cite{hu2017squeeze} \cite{cao2015look} \cite{wang2017residual}.

\section{Channel Locality Blocks}
\label{Channel Locality Blocks}
The Channel Locality block is a computational unit and the networks can be constructed by stacking convolutional layer with Channel Locality block. Each Channel Locality block is divided into tow parts: global information extraction part and nearby channel correlation extraction part. The global information extraction part performs feature compression in spatial dimensions, transforming each two-dimensional feature channel into real numbers. The real numbers have global receptive fields to some extent and have matched output dimensions as the input feature channels. The locality extraction part uses a single one-dimension filter to explicitly learn the correlation between nearby channel. Different from SEnet, we do not aim to learn the correlation between all channels. We believe that using the locality of channel helps loose the coupling between all channels. It will make the network more robust.

\subsection{Global Information Extraction}
\label{Global Information Extraction}
\begin{figure}[ht]
\vskip 0.2in
\begin{center}
\centerline{\includegraphics[width=\columnwidth]{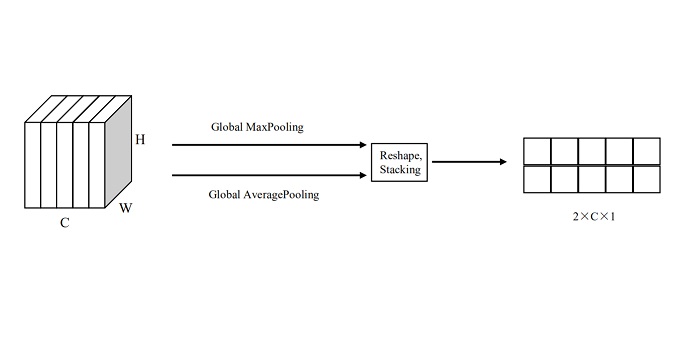}}
\caption{A diagram of the first step of a global information extraction part. We stack the global spatial information got by Global AveragePooling and Global Maxpooling.}
\label{fig2}
\end{center}
\vskip -0.2in
\end{figure}

\begin{figure}[ht]
\vskip 0.2in
\begin{center}
\centerline{\includegraphics[width=\columnwidth]{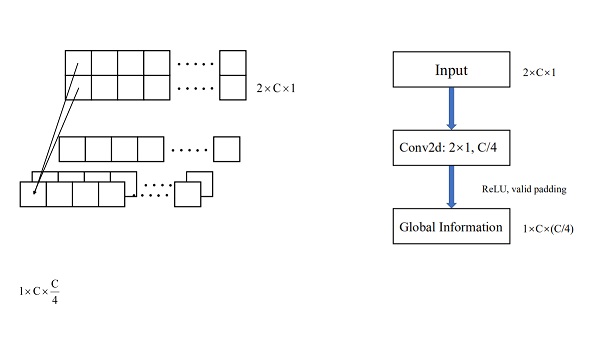}}
\caption{A diagram of the learning step of a global information extraction part. A group of $2 \times 1$ filters are applied to learn the relationship between two global information vectors.}
\label{fig3}
\end{center}
\vskip -0.2in
\end{figure}

Global spatial information is the most important value for describing each feature channel. There are usually two general methods for extracting global spatial information: (1) Global AveragePooling\cite{lin2013network}. (2) Global Maxpooling. In the squeeze operation of SEnet, the global spatial information is extracted by global average pooling. In this paper, we propose to combine the Global AveragePooling and Global Maxpooling to generate the channel descriptor. We use a group of weight layers to learn the linear combination of the global spatial information got by the two global pooling methods. We stack the global spatial information vectors to a two-dimension matrix as Figure~\ref{fig2} and then use a group of $2 \times 1$ filters to learn the relationship between the two vectors and calculate the linear combination of them. We adopt not only one filter, as the results of avoiding an incomplete of relationship information and trying to enable the model a stronger capability to learn the relationship. A diagram of a global information extraction part is shown in Figure~\ref{fig3}.

\subsection{Nearby Channel Correlation Extraction}
\label{Nearby Channel Correlation Extraction}
\begin{figure}[ht]
\vskip 0.2in
\begin{center}
\centerline{\includegraphics[width=\columnwidth]{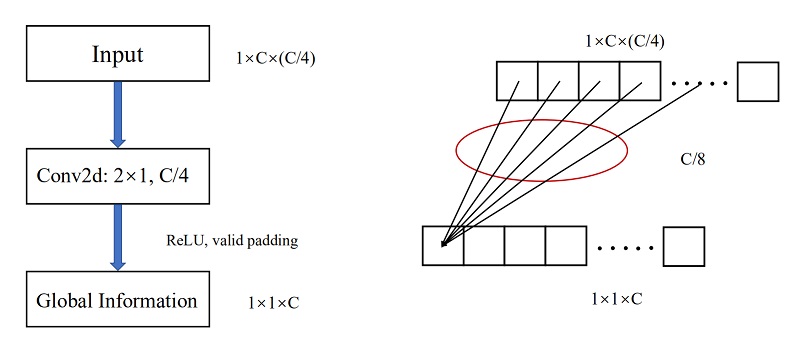}}
\caption{ A diagram of a nearby channel correlation extraction part. A single filter is applied to learn the correlation between the nearby channels.}
\label{fig4}
\end{center}
\vskip -0.2in
\end{figure}

To make use of the channel locality to generate channel attention for feature maps, we use a single filter as the weight layer to do a regression of a group of nearby channels. The filter learns the correlation between the nearby channels and the correlation between the channels and the empty fields at two ends of the strand. To fulfill the requirement of our excepted design, we set the length of the filters as $\frac{1}{4}$ as the length of the feature strand and adopt the same padding. Firstly, we believe that the different length of channel strands results in different characters of channel localities. The length of the filters which learns the nearby channel correlation should be adaptive to the length of the input channel strand length. On the other hand, learning the relationship between the channels and the empty fields at two ends of the strand helps the filter prepossessing the relationship between effective channels and inactive channels(whose values close to zeros). A diagram of a nearby channel correlation extraction part is shown in Figure~\ref{fig4}.

\section{Experiments}
\label{Experiments}
In this paper, we aim to compare the performance of our algorithm and SE block instead of pursuing state-of-art performance. Due to the limitation of computing resources and time, we perform experiments on CIFAR-10 \cite{krizhevsky2014cifar} datasets and evaluate our algorithm and SE block on different modified CNNs architectures which are small enough. Instead of getting higher accuracy on big datasets, we focus more on exploring the influence on CNNs performance between global channel correlation and local channel correlation. The models we used in our experiments were modified to suit the CIFAR-10 and CIFAR-100 dataset and trained with limited computing resources. The data augmentation method we adopt is horizontal flipping, and $0.2$ randomly shift. All the models are build by Keras \cite{chollet2015keras} framework. All experimental results of baselines are based on our actual training model.

For each experiment, we set the full connection layers of SE block with length $\frac{C}{8}$ and C, where C is the length of the input channel. The number of filters of C-Local block follows the principle we discuss in Chapter ~\ref{Global Information Extraction} and ~\ref{Nearby Channel Correlation Extraction}. We use Adam optimizer with initial learning rate $0.01$ and decay by multiplying $0.94$ for each two epoch. All test models are for $150$ epoch. Each convolutional layer is with a $0.0001$ $L2$ regularization and followed by a batch normalization layer \cite{ioffe2015batch}. We use He Normal as the initializer for the full connection layers of SE block. Through our experiments, if we merely use all ones or all zeros to initiate SE block, it will not improve the performance of CNNs.In some cases, it can even make the network perform very terrible. For controlling variables that will affect results in our experiments, we use the same initiation method for all C-Local blocks. All experimental results were averaged from five experiments.

\subsection{Experiments on Plane CNN}
We start our experiments by testing two channel attention method on a simple network. The model descriptions are shown as Table ~\ref{table1a}. The results are shown as Table ~\ref{table1b}.

\begin{table}[htbp]
\caption{Experiments on Plane CNN. (Left) plane CNN benchmark. (Middle) plane CNN with SE block. (Right) plane CNN with C-Local block.}
\label{table1a}
\vskip 0.15in
\begin{center}
\begin{tiny}
\begin{sc}
\begin{tabular}{ccccc}
\hline
\multicolumn{1}{|l|}{Output size} & \multicolumn{1}{l|}{Baseline}       & \multicolumn{1}{l|}{With SE block}                                                              & \multicolumn{1}{l|}{With C-Local block}                                                                     \\ \hline
\multicolumn{1}{|l|}{$32 \times 32$}       & \multicolumn{1}{l|}{conv, $3 \times 3$, 32}  & \multicolumn{1}{l|}{\begin{tabular}[c]{@{}l@{}}conv, $3 \times 3$, 32\\ fc, {[}4, 32{]}\end{tabular}}    & \multicolumn{1}{l|}{\begin{tabular}[c]{@{}l@{}}conv, $3 \times 3$, 32\\ conv, $2 \times 1$, 8\\ conv, $4 \times 1$, 1\end{tabular}}    \\ \hline
\multicolumn{1}{|l|}{$16 \times 16$}       & \multicolumn{3}{c|}{Max Pooling $2 \times 2$}                                                                                                                                                                                                                \\ \hline
\multicolumn{1}{|l|}{$16 \times 16$}       & \multicolumn{1}{l|}{conv, $3 \times 3$, 64}  & \multicolumn{1}{l|}{\begin{tabular}[c]{@{}l@{}}conv, $3 \times 3$, 64\\ fc, {[}16, 128{]}\end{tabular}}  & \multicolumn{1}{l|}{\begin{tabular}[c]{@{}l@{}}conv, $3 \times 3$, 64\\ conv, $2 \times 1$, 16\\ conv, $8 \times 1$, 1\end{tabular}}   \\ \hline
\multicolumn{1}{|l|}{$8 \times 8$}         & \multicolumn{3}{c|}{Max Pooling $2 \times 2$}                                                                                                                                                                                                                \\ \hline
\multicolumn{1}{|l|}{$8 \times 8$}         & \multicolumn{1}{l|}{conv, $3 \times 3$, 128} & \multicolumn{1}{l|}{\begin{tabular}[c]{@{}l@{}}conv, $3 \times 3$, 128\\ fc, {[}16, 128{]}\end{tabular}} & \multicolumn{1}{l|}{\begin{tabular}[c]{@{}l@{}}conv, $3 \times 3$, 128\\ conv, $2 \times 1$, 32\\ conv, $16 \times 1$, 1\end{tabular}} \\ \hline
\multicolumn{1}{|l|}{$8 \times 8$}         & \multicolumn{1}{l|}{conv, $3 \times 3$, 128} & \multicolumn{1}{l|}{\begin{tabular}[c]{@{}l@{}}conv, $3 \times 3$, 128\\ fc, {[}16, 128{]}\end{tabular}} & \multicolumn{1}{l|}{\begin{tabular}[c]{@{}l@{}}conv, $3 \times 3$, 128\\ conv, $2 \times 1$, 32\\ conv, $16 \times 1$, 1\end{tabular}} \\ \hline
$1 \times 1$                               & \multicolumn{3}{c}{global average pool, 10-d fc, softmax}
\end{tabular}
\end{sc}
\end{tiny}
\end{center}
\vskip -0.1in
\end{table}

\begin{table}[htbp]
\caption{Experiments on Plane CNN. Accuracy ($\%$) the cifar-10 test set.}
\label{table1b}
\vskip 0.15in
\begin{center}
\begin{small}
\begin{sc}
\begin{tabular}{lcccr}
\toprule
 Model & cifar-10 accuracy \\
\midrule
Baseline    & 84.5\\
With SE block & 85.6 \\
With V-Local block &86.8 \\
\bottomrule
\end{tabular}
\end{sc}
\end{small}
\end{center}
\vskip -0.1in
\end{table}

\subsection{Experiments on ALL CNN}
In this experiment, we apply C-Local block and SE block on ALL CNN \cite{springenberg2014striving}. The model descriptions are shown as Table ~\ref{table2a}. The results are shown as Table ~\ref{table2b}.

\begin{table}[htbp]
\caption{Experiments on ALL CNN. (Left) plane CNN benchmark. (Middle) plane CNN with SE block. (Right) plane CNN with C-Local block.}
\label{table2a}
\vskip 0.15in
\begin{center}
\begin{tiny}
\begin{sc}
\begin{tabular}{lcccr}
\hline
Output size & Baseline                                                           & With SE block                                                                      & With C-Local block                                                                                 \\ \hline
$32 \times 32$       & conv, $3 \times 3$, 64                                                      & \begin{tabular}[c]{@{}l@{}}conv, $3 \times 3$, 64\\ fc, [8, 64]\end{tabular}                & \begin{tabular}[c]{@{}l@{}}conv, $3 \times 3$, 64\\ conv, $2 \times 1$, 16\\ conv, $8 \times 1$, 1\end{tabular}               \\ \hline
$32 \times 32$       & conv, $3 \times 3$, 64                                                      & \begin{tabular}[c]{@{}l@{}}conv, $3 \times 3$, 64\\ fc, [8, 64]\end{tabular}                & \begin{tabular}[c]{@{}l@{}}conv, $3 \times 3$, 64\\ conv, $2 \times 1$, 16\\ conv, $8 \times 1$, 1\end{tabular}               \\ \hline
$16 \times 16$       & \begin{tabular}[c]{@{}l@{}}conv, $3 \times 3$, 64\\ Strides:2\end{tabular}  & \begin{tabular}[c]{@{}l@{}}conv, $3 \times 3$, 64\\ Strides:2\\ fc, [8, 64]\end{tabular}    & \begin{tabular}[c]{@{}l@{}}conv, $3 \times 3$, 64\\ Strides:2\\ conv, $2 \times 1$, 16\\ conv, $8 \times 1$, 1\end{tabular}   \\ \hline
$16 \times 16$       & \multicolumn{1}{c|}{conv, $3 \times 3$, 128}                                & \begin{tabular}[c]{@{}l@{}}conv, $3 \times 3$, 128\\ fc, [16, 128]\end{tabular}             & \begin{tabular}[c]{@{}l@{}}conv, $3 \times 3$, 128\\ conv, $2 \times 1$, 32\\ conv, $16 \times 1$, 1\end{tabular}             \\ \hline
$16 \times 16$       & conv, $3 \times 3$, 128                                                     & \begin{tabular}[c]{@{}l@{}}conv, $3 \times 3$, 128\\ fc, [16, 128]\end{tabular}             & \begin{tabular}[c]{@{}l@{}}conv, $3 \times 3$, 128\\ conv, $2 \times 1$, 32\\ conv, $16 \times 1$, 1\end{tabular}             \\ \hline
$8 \times 8$         & \begin{tabular}[c]{@{}l@{}}conv, $3 \times 3$, 128\\ Strides:2\end{tabular} & \begin{tabular}[c]{@{}l@{}}conv, $3 \times 3$, 128\\ Strides:2\\ fc, [16, 128]\end{tabular} & \begin{tabular}[c]{@{}l@{}}conv, $3 \times 3$, 128\\ Strides:2\\ conv, $2 \times 1$, 32\\ conv, $16 \times 1$, 1\end{tabular} \\ \hline
$8 \times 8$         & conv, $1 \times 1$, 128                                                     & \begin{tabular}[c]{@{}l@{}}conv, $1 \times 1$, 128\\ fc, [16, 128]\end{tabular}             & \begin{tabular}[c]{@{}l@{}}conv, $1 \times 1$, 128\\ conv, $2 \times 1$, 32\\ conv, $16 \times 1$, 1\end{tabular}             \\ \hline
$8 \times 8$         & \multicolumn{3}{c|}{conv, $1 \times 1$, 10}                                                                                                                                                                                                                           \\ \hline
$1 \times 1$         & \multicolumn{3}{c|}{global average pool, softmax}                                                                                                                                                                                                            \\ \hline
\end{tabular}
\end{sc}
\end{tiny}
\end{center}
\vskip -0.1in
\end{table}

\begin{table}[htbp]
\caption{Experiments on ALL CNN. Accuracy ($\%$) the cifar-10 test set.}
\label{table2b}
\vskip 0.15in
\begin{center}
\begin{small}
\begin{sc}
\begin{tabular}{lcccr}
\toprule
 Model & cifar-10 accuracy \\
\midrule
Baseline    & 88.7\\
With SE block & 89.4 \\
With V-Local block &90.8 \\
\bottomrule
\end{tabular}
\end{sc}
\end{small}
\end{center}
\vskip -0.1in
\end{table}

\subsection{ Experiments on modified ResNet}
\begin{figure}[htbp]
\vskip 0.2in
\begin{center}
\centerline{\includegraphics[width=\columnwidth]{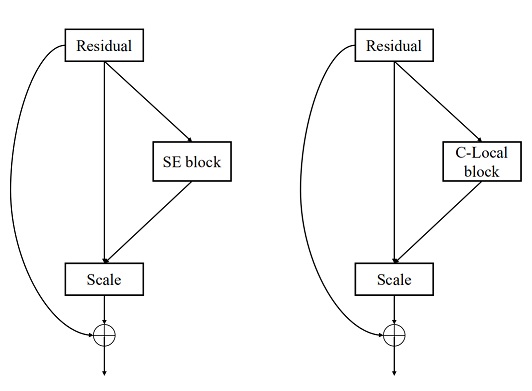}}
\caption{ A diagram to show how we apply SE block and C-Local block to ResNet.}
\label{fig5}
\end{center}
\vskip -0.2in
\end{figure}

In this experiment, we apply C-Local block and SE block on a modified ResNet \cite{he2016deep}. The model descriptions are shown as Table ~\ref{table3a}. We follow the design principle in the original paper \cite{hu2017squeeze} of SE block to apply SE block and C-Local block. The diagram is shown in Figure ~\ref{fig5}. The results are shown as Table ~\ref{table3b}.

\begin{table}[htbp]
\caption{Experiments on modified ResNet. (Left) plane CNN benchmark. (Middle) plane CNN with SE block. (Right) plane CNN with C-Local block.}
\label{table3a}
\vskip 0.15in
\begin{center}
\begin{tiny}
\begin{sc}
\begin{tabular}{lcccr}
\hline
Output size & Baseline                                                                               & With SE block                                                                                          & With C-Local block                                                                                                        \\ \hline
$32 \times 32$      & \multicolumn{3}{c|}{conv, $3 \times 3$, 32}                                                                                                                                                                                                                                                                                          \\ \hline
$32 \times 32$      & \begin{tabular}[c]{@{}l@{}}conv, $1 \times 1$, 16\\ conv, $3 \times 3$, 16\\ conv, $1 \times 1$, 32\end{tabular}  & \begin{tabular}[c]{@{}l@{}}conv, $1 \times 1$, 16\\ conv, $3 \times 3$, 16\\ conv, $1 \times 1$, 32\\ fc, [4, 32]\end{tabular}    & \begin{tabular}[c]{@{}l@{}}conv, $1 \times 1$, 16\\ conv, $3 \times 3$, 16\\ conv, $1 \times 1$, 32\\ \\ \\ conv, $2 \times 1$, 8\\ conv, $4 \times 1$, 1\end{tabular} \\ \hline
$16 \times 16$      & \multicolumn{3}{c|}{MaxPooling $2 \times 2$}                                                                                                                                                                                                                                                                                         \\ \hline
$16 \times 16$      & \begin{tabular}[c]{@{}l@{}}conv, $1 \times 1$, 32\\ conv, $3 \times 3$, 32\\ conv, $1 \times 1$, 64\end{tabular}  & \begin{tabular}[c]{@{}l@{}}conv, $1 \times 1$, 32\\ conv, $3 \times 3$, 32\\ conv, $1 \times 1$, 64\\ fc, [8, 64]\end{tabular}    & \begin{tabular}[c]{@{}l@{}}conv, $1 \times 1$, 32\\ conv, $3 \times 3$, 32\\ conv, $1 \times 1$, 64\\ conv, $2 \times 1$, 16\\ conv, $8 \times 1$, 1\end{tabular}      \\ \hline
$8 \times 8$         & \multicolumn{3}{c|}{MaxPooling $2 \times 2$}                                                                                                                                                                                                                                                                                         \\ \hline
$8 \times 8$         & \begin{tabular}[c]{@{}l@{}}conv, $1 \times 1$, 64\\ conv, $3 \times 3$, 64\\ conv, $1 \times 1$, 128\end{tabular} & \begin{tabular}[c]{@{}l@{}}conv, $1 \times 1$, 64\\ conv, $3 \times 3$, 64\\ conv, $1 \times 1$, 128\\ fc, [16, 128]\end{tabular} & \begin{tabular}[c]{@{}l@{}}conv, $1 \times 1$, 64\\ conv, $3 \times 3$, 64\\ conv, $1 \times 1$, 128\\ conv, $2 \times 1$, 32\\ conv, $16 \times 1$, 1\end{tabular}    \\ \hline
$8 \times 8$         & \begin{tabular}[c]{@{}l@{}}conv, $1 \times 1$, 64\\ conv, $3 \times 3$, 64\\ conv, $1 \times 1$, 128\end{tabular} & \begin{tabular}[c]{@{}l@{}}conv, $1 \times 1$, 64\\ conv, $3 \times 3$, 64\\ conv, $1 \times 1$, 128\\ fc, [16, 128]\end{tabular} & \begin{tabular}[c]{@{}l@{}}conv, $1 \times 1$, 64\\ conv, $3 \times 3$, 64\\ conv, $1 \times 1$, 128\\ conv, $2 \times 1$, 32\\ conv, $16 \times 1$, 1\end{tabular}    \\ \hline
            & \multicolumn{3}{c|}{global average pool, 10-d fc, softmax}                                                                                                                                                                                                                                                                  \\ \hline
\end{tabular}
\end{sc}
\end{tiny}
\end{center}
\vskip -0.1in
\end{table}

\begin{table}[htbp]
\caption{Experiments on modified ResNet. Accuracy ($\%$) the cifar-10 test set.}
\label{table3b}
\vskip 0.15in
\begin{center}
\begin{small}
\begin{sc}
\begin{tabular}{lcccr}
\toprule
 Model & cifar-10 accuracy \\
\midrule
Baseline    & 86.7\\
With SE block & 87.4 \\
With V-Local block &88.3 \\
\bottomrule
\end{tabular}
\end{sc}
\end{small}
\end{center}
\vskip -0.1in
\end{table}

\section{Future work}
In this paper, we did not evaluate C-Local block on the bigger dataset. After we get enough computing resources, we will try to apply our algorithm to ImageNet and other large datasets. At the same time, we think our current algorithm is rough. In the future work, we will optimize and improve our algorithm as much as possible.

\section{Conclusion}
In this paper we proposed the C-Local block, a variant of SE block designed to improve the representational capacity of a network by enabling it to perform dynamic channel-wise feature recalibration. Extensive experiments demonstrate the effectiveness of C-Local block which achieve better performance on cifar-10 datasets when we apply these two blocks to different CNNs architectures. Although we only validated our algorithm on a small dataset, we still have reason to believe that our algorithm has the potential to be an excellent algorithm like SE. Finally,through our C-Local block and previous work on CNNs,exploring the relationship between local data is probably a better choice than exploring global data.

\nocite{langley00}

\bibliography{citation}

\begin{thebibliography}{10}
\providecommand{\url}[1]{#1}
\csname url@samestyle\endcsname
\providecommand{\newblock}{\relax}
\providecommand{\bibinfo}[2]{#2}
\providecommand{\BIBentrySTDinterwordspacing}{\spaceskip=0pt\relax}
\providecommand{\BIBentryALTinterwordstretchfactor}{4}
\providecommand{\BIBentryALTinterwordspacing}{\spaceskip=\fontdimen2\font plus
\BIBentryALTinterwordstretchfactor\fontdimen3\font minus
  \fontdimen4\font\relax}
\providecommand{\BIBforeignlanguage}[2]{{%
\expandafter\ifx\csname l@#1\endcsname\relax
\typeout{** WARNING: IEEEtran.bst: No hyphenation pattern has been}%
\typeout{** loaded for the language `#1'. Using the pattern for}%
\typeout{** the default language instead.}%
\else
\language=\csname l@#1\endcsname
\fi
#2}}
\providecommand{\BIBdecl}{\relax}
\BIBdecl

\bibitem{hu2017squeeze}
J.~Hu, L.~Shen, and G.~Sun, ``Squeeze-and-excitation networks,'' \emph{arXiv
  preprint arXiv:1709.01507}, vol.~7, 2017.

\bibitem{krizhevsky2012imagenet}
A.~Krizhevsky, I.~Sutskever, and G.~E. Hinton, ``Imagenet classification with
  deep convolutional neural networks,'' in \emph{Advances in neural information
  processing systems}, 2012, pp. 1097--1105.

\bibitem{ren2015faster}
S.~Ren, K.~He, R.~Girshick, and J.~Sun, ``Faster r-cnn: Towards real-time
  object detection with region proposal networks,'' in \emph{Advances in neural
  information processing systems}, 2015, pp. 91--99.

\bibitem{long2015fully}
J.~Long, E.~Shelhamer, and T.~Darrell, ``Fully convolutional networks for
  semantic segmentation,'' in \emph{Proceedings of the IEEE conference on
  computer vision and pattern recognition}, 2015, pp. 3431--3440.

\bibitem{xie2017aggregated}
S.~Xie, R.~Girshick, P.~Doll{\'a}r, Z.~Tu, and K.~He, ``Aggregated residual
  transformations for deep neural networks,'' in \emph{Computer Vision and
  Pattern Recognition (CVPR), 2017 IEEE Conference on}.\hskip 1em plus 0.5em
  minus 0.4em\relax IEEE, 2017, pp. 5987--5995.

\bibitem{szegedy2015going}
C.~Szegedy, W.~Liu, Y.~Jia, P.~Sermanet, S.~Reed, D.~Anguelov, D.~Erhan,
  V.~Vanhoucke, and A.~Rabinovich, ``Going deeper with convolutions,'' in
  \emph{Proceedings of the IEEE conference on computer vision and pattern
  recognition}, 2015, pp. 1--9.

\bibitem{jaderberg2015spatial}
M.~Jaderberg, K.~Simonyan, A.~Zisserman \emph{et~al.}, ``Spatial transformer
  networks,'' in \emph{Advances in neural information processing systems},
  2015, pp. 2017--2025.

\bibitem{lecun1995convolutional}
Y.~LeCun, Y.~Bengio \emph{et~al.}, ``Convolutional networks for images, speech,
  and time series,'' \emph{The handbook of brain theory and neural networks},
  vol. 3361, no.~10, p. 1995, 1995.

\bibitem{ma2018shufflenet}
N.~Ma, X.~Zhang, H.-T. Zheng, and J.~Sun, ``Shufflenet v2: Practical guidelines
  for efficient cnn architecture design,'' \emph{arXiv preprint
  arXiv:1807.11164}, vol.~5, 2018.

\bibitem{chollet2017xception}
F.~Chollet, ``Xception: Deep learning with depthwise separable convolutions,''
  \emph{arXiv preprint}, pp. 1610--02\,357, 2017.

\bibitem{cao2015look}
C.~Cao, X.~Liu, Y.~Yang, Y.~Yu, J.~Wang, Z.~Wang, Y.~Huang, L.~Wang, C.~Huang,
  W.~Xu \emph{et~al.}, ``Look and think twice: Capturing top-down visual
  attention with feedback convolutional neural networks,'' in \emph{Proceedings
  of the IEEE International Conference on Computer Vision}, 2015, pp.
  2956--2964.

\bibitem{wang2017residual}
F.~Wang, M.~Jiang, C.~Qian, S.~Yang, C.~Li, H.~Zhang, X.~Wang, and X.~Tang,
  ``Residual attention network for image classification,'' \emph{arXiv preprint
  arXiv:1704.06904}, 2017.

\bibitem{lin2013network}
M.~Lin, Q.~Chen, and S.~Yan, ``Network in network,'' \emph{arXiv preprint
  arXiv:1312.4400}, 2013.

\bibitem{krizhevsky2014cifar}
A.~Krizhevsky, V.~Nair, and G.~Hinton, ``The cifar-10 dataset,'' \emph{online:
  http://www. cs. toronto. edu/kriz/cifar. html}, 2014.

\bibitem{chollet2015keras}
F.~Chollet \emph{et~al.}, ``Keras,'' 2015.

\bibitem{ioffe2015batch}
S.~Ioffe and C.~Szegedy, ``Batch normalization: Accelerating deep network
  training by reducing internal covariate shift,'' \emph{arXiv preprint
  arXiv:1502.03167}, 2015.

\bibitem{springenberg2014striving}
J.~T. Springenberg, A.~Dosovitskiy, T.~Brox, and M.~Riedmiller, ``Striving for
  simplicity: The all convolutional net,'' \emph{arXiv preprint
  arXiv:1412.6806}, 2014.

\bibitem{he2016deep}
K.~He, X.~Zhang, S.~Ren, and J.~Sun, ``Deep residual learning for image
  recognition,'' in \emph{Proceedings of the IEEE conference on computer vision
  and pattern recognition}, 2016, pp. 770--778.

\end{thebibliography}
\bibliographystyle{IEEEtran}

\end{document}